\documentclass[10pt,letterpaper]{article}

\usepackage{cogsci}

\cogscifinalcopy % Uncomment this line for the final submission 

\usepackage{pslatex}
\usepackage{apacite}
\usepackage{float} % Roger Levy added this and changed figure/table
\usepackage{graphicx}
\usepackage{amsmath}
\usepackage{amsfonts}
\newcommand{\code}[1]{\texttt{#1}}
\newcommand{\qtext}[1]{\textit{``#1''}}

\DeclareMathOperator*{\argmin}{arg\,min}

\usepackage[T1]{fontenc}
\usepackage{inconsolata}
\newcommand{\church}[1]{\code{#1}}

\title{
Evaluating statistical language models as pragmatic reasoners
}

\author{
Benjamin Lipkin$^{1}$\quad Lionel Wong$^{1}$\quad Gabriel Grand$^{2}$\quad Joshua B. Tenenbaum$^{1,2}$\\
$^1$BCS, MIT\quad$^2$CSAIL, MIT\\
\code{\{lipkinb, zyzzyva, gg, jbt\}@mit.edu}
}

\begin{document}

\maketitle

\begin{abstract}

The relationship between communicated language and intended meaning is often probabilistic and sensitive to context.
Numerous strategies attempt to estimate such a mapping, often leveraging recursive Bayesian models of communication.
In parallel, large language models (LLMs) have been increasingly applied to semantic parsing applications, tasked with inferring logical representations from natural language.
While existing LLM explorations have been largely restricted to literal language use, in this work, we evaluate the capacity of LLMs to infer the meanings of pragmatic utterances. 
Specifically, we explore the case of threshold estimation on the gradable adjective \qtext{strong}, contextually conditioned on a strength prior, then extended to composition with qualification, negation, polarity inversion, and class comparison.
We find that LLMs can derive context-grounded, human-like distributions over the interpretations of several complex pragmatic utterances, yet struggle composing with negation.
These results inform the inferential capacity of statistical language models, and their use in pragmatic and semantic parsing applications.
All corresponding code is made publicly available\footnote{https://github.com/benlipkin/probsem/tree/CogSci2023}.


\end{abstract}

\section{Introduction}

Natural language understanding unfolds in context and reflects more than literal interpretation.
Such a process is posited to be mediated by a series of inferences, which jointly scrutinize mappings between linguistic structure and mental representations in tandem with the plausibility of resulting interpretations.
A sentence as simple as \qtext{Mia is tall} may be broadly meaningful in of itself, but the range of plausible heights a listener will consider shifts with context that \qtext{Mia plays in the WNBA} or that \qtext{Mia is a three-year old child.} These contextual inferences are broadly studied as linguistic \textit{pragmatics} \cite{wittgenstein2010philosophical,searle1969speech,austin1975things,levinson1983pragmatics,grice1989studies,clark1996using}.

Recently, work on large-scale training of transformer language models has produced engineering artifacts that perform exceedingly well across a range of natural language processing (NLP) benchmarks.
While trained explicitly to optimize an objective of next-token prediction, such systems implicitly recapitulate large swaths of the traditional NLP pipeline, from POS tagging and parsing to semantic role labeling and coreference resolution \cite{tenney2019bert,bommasani2021opportunities}.
Indeed, a growing body of contemporary work utilizes LLMs to synthesize program-like representations from natural language (NL) inputs for use in downstream applications from action planning to theorem solving \cite{acquaviva2021communicating,gao2022pal,collins2022structured,mishra2022lila,zelikman2022parsel,wong2023translating}. 
In leveraging such systems as semantic parsers, this work casts LLMs as formal accounts of the mapping between linguistic forms and representations of meaning. 
However, such evaluations have been largely restricted to \textit{literal} language use and translation.
In contrast, \textit{pragmatic} meaning estimation often requires considering a distribution over multiple interpretations in context, presenting additional complexity \cite{fried2022pragmatics,hu2022fine,ruis2022large,hu2023expectations}.

Existing models of pragmatic reasoning typically rely on explicit probabilistic computation, often within the \textit{Rational Speech Acts} (RSA) communication framework, whereby a pragmatic listener reasons about an informative speaker to infer intended meanings \cite{frank2012predicting,goodman2013knowledge,goodman2016pragmatic}. We ask: can statistical language models \textit{amortize} common pragmatic inferences, recovering approximately equivalent distributions between language and contextually-modulated meanings?

To address this question, in this work, we explore the case of interpretation over the gradable adjective \qtext{strong} in describing a player in a fictional game.
Conditioned on context describing a generative model over possible worlds, expressing a numerical prior on \qtext{strength}, among other variables, our paradigm invokes estimation over numerical interpretations of textual descriptions of a novel player's strength.
We collect both LLM-estimated and human-measured distributions over the interpretations of such utterances, and explore composition with additional dimensions of complexity.
We find that LLMs impressively infer context-aware, human-like distributions over complex pragmatic utterances such as \qtext{very strong for a beginner}.
Simultaneously, we observe a failure to compose such inferred meanings with negation, e.g., \qtext{not strong} or polarity inversion, e.g., \qtext{weak}, offering insights into potential shortcomings.

\subsection{Meaning as probabilistic programs}

In expressing formal representations of linguistic meaning, one approach has been to build from the framework of model theoretic semantics \cite{kripke1963semantical,montague1973proper,partee1990mathematical,kratzer1998semantics}, in combination with uncertainty quantification \cite{van2012probabilistic,cooper2015probabilistic}, converging upon probabilistic programming languages (PPLs), like Church \cite{goodman2012church}, as a useful substrate.
\citeauthor{goodman2015probabilistic} \citeyear{goodman2015probabilistic}, in particular, present a framework, which we build from here, for NL as belief updating over probabilistic programs.
Starting from a generative model over possible worlds describing a domain, sentences are incrementally expressed as conditioning statements and executed to update posterior beliefs over world states.

\citeauthor{goodman2015probabilistic} motivate this framework by providing examples through a discussion of a fictional game of tug-of-war (ToW).
In this simplified version of the classical children’s game, two teams, each with one or more players, compete against each other, with the winner decided by the team whose players exert the most strength \cite{probmods,gerstenberg2012ping,goodman2014concepts}.
Starting from this base, \citeauthor{goodman2015probabilistic} built examples of PPL-mediated contextual semantic analysis.
For example, \qtext{Team A has more than 3 players} could be expressed as \church{(condition (> (length team-a) 3))}, and when queried if \qtext{Team A} might beat \qtext{Team B} (which perhaps has only 2 players), this information would be considered in evaluating the distribution over outcomes of such a match.
In elevating this approach beyond literal language use, to scenarios where NL presents with nondeterministic interpretation, \citeauthor{goodman2015probabilistic} proposed leveraging explicit probabilistic computation via RSA.
One difficulty with this framework is the need to manually synthesize programs expressing the semantics of evaluated NL.
Drawing from successful approaches in semantic parsing and program synthesis, such a process lends itself increasingly to automation using LLMs.

\subsection{Present study}

\begin{figure}[h]
\begin{center}
\includegraphics[width=0.49\textwidth]{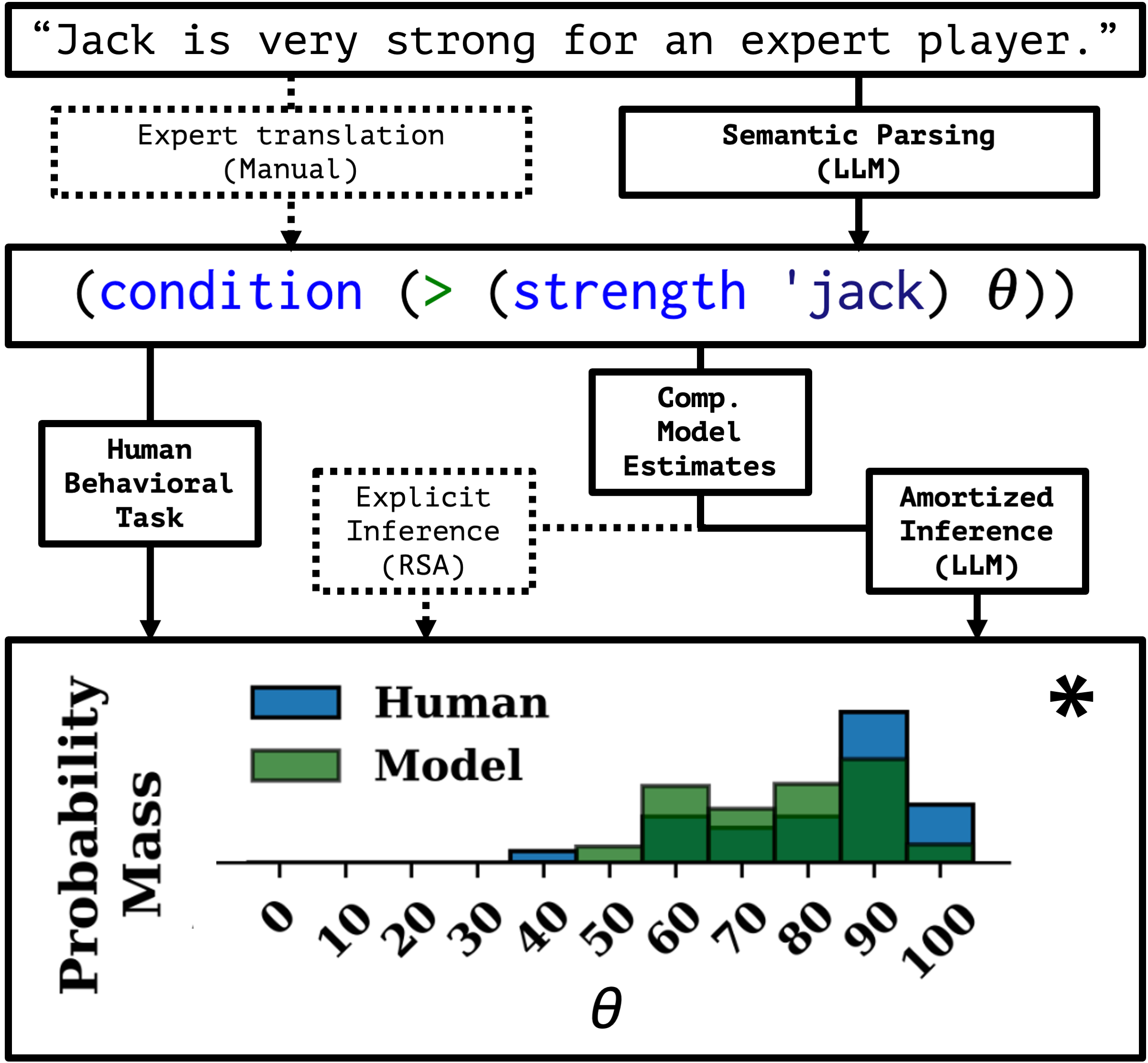}
\end{center}
\caption{Schematic overview. LLMs stand in for the traditional pragmatics pipeline, often recovering human-like estimates over multiple interpretations of complex constructions.} 
\label{overview}
\end{figure}

\citeauthor{goodman2015probabilistic} \citeyear{goodman2015probabilistic} have highlighted the elegant capacity of PPLs in expressing the logical representation of sentence meaning, but have left open how such programs be derived in the first place.
In parallel, modern semantic parsing work has painted a picture of LLMs as systems capable of mediating such a translation.
However, when it comes to scenarios where this task moves beyond literal language use, it is unclear: a) if LLMs are appropriately suited to mediate such sophisticated inferences and b), whether such model estimates would be in line with human expectations.
In addressing these questions, we build from the ToW domain model and pursue gradable adjectives as an expressive test bed.

Gradable adjectives, such as \qtext{strong}, present with vagueness as they lack precise class boundaries.
Several approaches have been developed to express the semantics of gradable adjectives, and in one common approach, a free threshold variable is introduced such that \qtext{strong} be defined as having \qtext{strength} $> \theta$ \cite{cresswell1976semantics,klein1980semantics,kennedy2007vagueness,lassiter2017adjectival,tessler2020informational}.
While the distribution over $\theta$ or other formulations can be derived to various degrees using the recursive probabilistic inference of RSA \cite{qing2014gradable,tessler2022warm}, here we ask whether an LLM can stand in, directly estimating the distribution over $\theta$ in a single forward pass. See Figure \ref{overview} for an overview.

Within the context of ToW players, with a prior over \qtext{strength} defined in the domain description, we begin with the basic evaluation of \qtext{strong} and its inverse polarity counterpart \qtext{weak}, then extending to the inclusion of negation, e.g., \qtext{not strong}, qualifiers, e.g., \qtext{pretty strong}, and comparison classes, e.g., \qtext{strong for a novice player}.
In considering the plausibility of LLM inferences, we collect human behavioral norms for the same stimuli to quantify where and how the model captures or fails to reflect human intuitions.
We find, on the positive end, that LLMs can perform rather sophisticated contextual amortization of a stack of inferences that include both literal and pragmatic ones, elegantly parsing over complex pragmatic utterances, conditioned on text expressing a generative world model as a probabilistic program.
On the negative end, however, LLMs can struggle with the otherwise logically simpler properties of negation or polarity inversion, deviating from human interpretations in such cases.
These results inform our understanding of the inferential capacity of LLMs, and as such simultaneously inform debates surrounding the capacities of statistical language learners (see e.g., \citeauthor{piantadosi2023modern} \citeyear{piantadosi2023modern}).

\section{Methods}

To explore the questions outlined thus far, we begin by more formally defining the ToW domain model in Church, outlining the priors and constraints placed on the semantics explored for the remainder of this work.
We define a scoring function, by which the LLM can estimate the probability of particular text interpretations, conditioned on the domain model context and an NL query.
To evaluate the efficacy of this scoring function, we developed a set of test materials to evaluate human and LLM-based interpretations of gradable adjectives, and tested our modeling framework and 60 human participants on two variations of this task, one focused primarily on qualification and one on class comparison.
Negation and polarity inversion were also explored as part of the qualification experiment.
Finally, we consider the distributions over interpretations estimated by the model with respect to those empirically measured in human participants.

\subsection{Domain Model and LLM Context}

\begin{figure}[h]
\begin{center}
\includegraphics[width=0.49\textwidth]{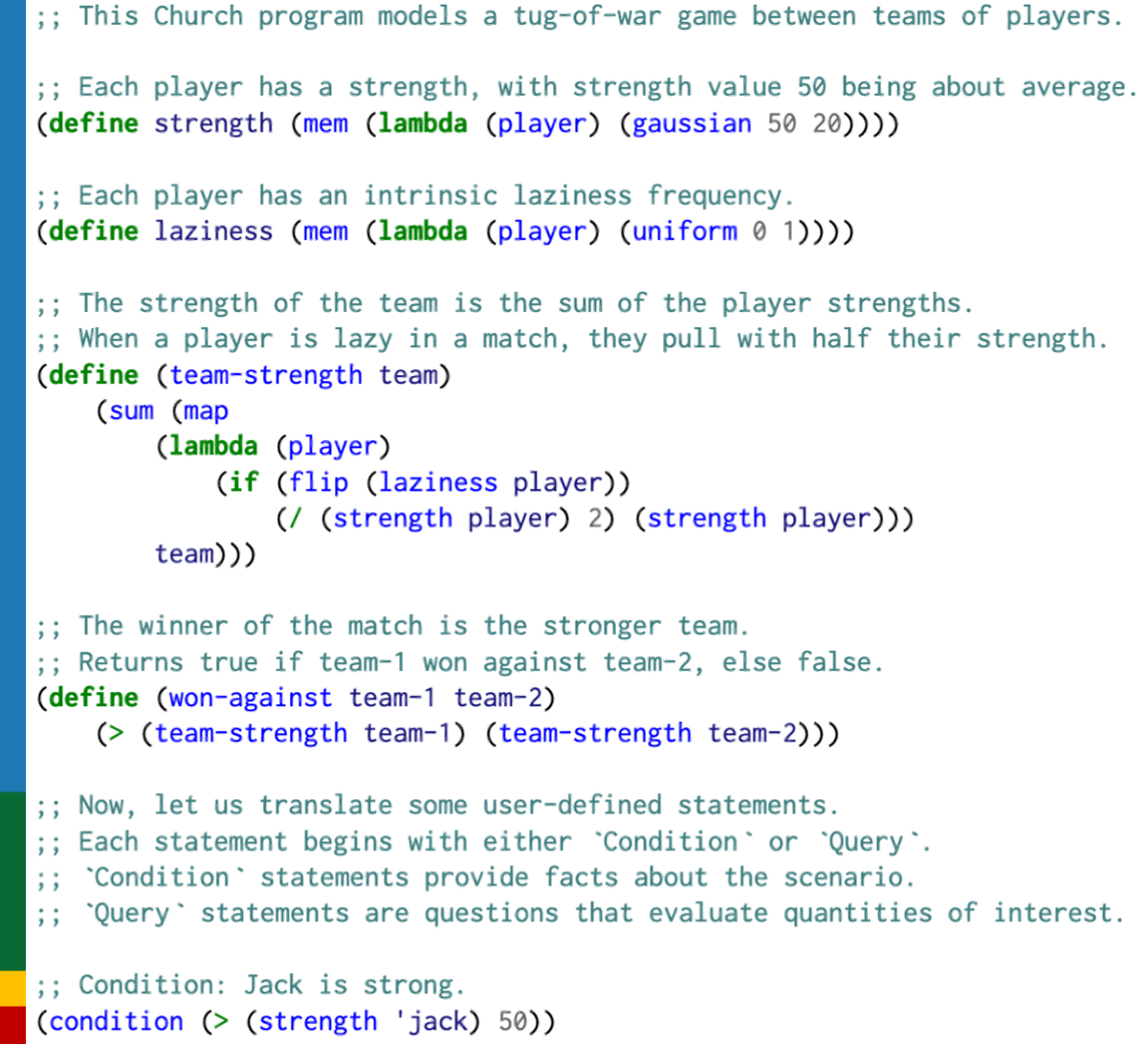}
\end{center}
\caption{Example of full text passed to LLM for a single query. The tug-of-war domain model (blue) and task instructions (green) are consistent across all trials. For each evaluated sentence (yellow), the probability of each program (red) is evaluated to return a score for a given interpretation.} 
\label{prompt}
\end{figure}

As the context for our gradable adjective experiments, we consider the previously introduced domain of tug-of-war.
In Figure \ref{prompt}, we present the precise formulation of this domain as a Church program.
Conditioned on this context describing the domain, a description of the task at hand (to translate NL into Church), and a particular NL query, an LLM can then act as a generative model over program expressions, with the capacity both to sample next tokens starting from the prompt, or to assign probabilities to predefined programs under the model.
Critically, we see a prior over \church{strength} $\sim \mathcal{N}(50,20)$.
While not all other elements of this domain model are required for our downstream tasks, we include full context so as to evaluate efficacy and robustness within a complete world model.

\subsection{Task Description}

To condition onto our world model that \qtext{Jack is strong}, expressed as \church{(condition (> (strength 'jack) $\theta$))}, what value for $\theta$ is reasonable?
While leveraging RSA is one strategy, it quickly grows intractable to accurately estimate such a range for all gradable adjectives, with the combinatorial space further plagued by the possible composition with additional constraints, e.g., \qtext{somewhat strong}.
So we ask: can an LLM amortize inference of this distribution over $\theta$ in a way that is pragmatically-sensitive and consistent with human inferences?
To evaluate this question, we developed a set of stimuli, each referencing gradable adjectives to describe the strength of a fictional athlete named \qtext{Jack}.
These materials were divided among two experiments.

\subsubsection{E1: Qualifiers}

In E1, we first evaluated the probability of sentences about Jack's strength to be interpreted as programs of the form: \church{(condition (> (strength 'jack) $\theta$))}, for $\theta$ from $0-100$, in intervals of $10$.
These included both cases where Jack is \qtext{strong} and where he is \qtext{not weak}, to various degrees.
For each sentence, $P_{model}(\theta)$ was estimated by the LLM, and $P_{human}(\theta)$ was measured from a collection of human participant point estimates.
In addition to these test sentences, a control sentence was included of the form, \qtext{Jack has at least average strength}, which lacks vagueness and has intention of recovering the majority of probability mass at $\theta = \mu_{strength} = 50$.
Then, to test robustness to polarity inversion, we developed a parallel set of materials to evaluate Jack's weakness, considering instead programs of the form \church{(condition (< (strength 'jack) $\theta$))}.
%[\textit{Note: The variant \church{(condition (not (> (strength 'jack) |$\theta$|)))} was also considered and results in identical conclusions.}]
These materials were directly matched to those in the first part of E1, with only the modification of swapping \qtext{strong} and \qtext{not weak} to \qtext{not strong} and \qtext{weak}, respectively.
The full set of 18 materials can be found in Figure \ref{av1-plot}.

\subsubsection{E2: Comparison Classes}

In E2, we extended this evaluation by introducing comparison classes to conditionally refine interpretation.
We modified the definition of \church{strength} in the LLM prompt, to consider a new variable, the \church{league} of a player, by injecting the following conditional statement (the full updated prompt can be found in the paper repository):\\
\church{
(cond\\
  \indent ((equal? league 'beginner)\\
  \indent (gaussian 30 20))\\
  \indent ((equal? league 'intermediate)\\
  \indent (gaussian 50 20))\\
  \indent ((equal? league 'professional)\\
  \indent (gaussian 70 20))\\
)
}\\
Here we test, can an LLM use a verbal descriptor of a player to jointly infer their league membership as well as relative strength within that league?
Drawing from a subset of E1, we developed a new set of materials that incorporate these comparison classes.
In particular, we preserved the control form: \qtext{Jack has at least average strength} and the form which deviated most from the mean in Figure \ref{av1-plot}: \qtext{Jack is very strong}.
We modified each sentence for each league, along three degrees of abstraction: exact match, synonym, and allusion.
For example, for the first league, we assessed Jack's strength for a \qtext{beginner}, \qtext{novice}, and \qtext{someone new to the game.}
The full set of 18 materials can be found in Figure \ref{av2-plot}.

\subsection{Human Participant Evaluation}

In order to evaluate $P_{human}(\theta)$ for each stimulus, two behavioral studies were conducted.
60 participants were recruited from Prolific, 30 for E1 and 30 for E2.
Participants provided informed
consent and were paid approximately \$15 per hour.
The experiment requested that participants move a slider to indicate the threshold ($\theta$) on the strength of a fictional athlete named \qtext{Jack}, based on independent readings of the stimulus sentences.
One participant was removed from E1 for self-reported comprehension difficulties.
Analyses include only the remaining participants.
The experimental source files, including instructions and stimulus materials, are released with the paper repository.

\subsection{LLM Scoring Function}

In order to evaluate $P_{model}(\theta)$ for each stimulus, a scoring function was defined over programs varying $\theta$.
The OpenAI \code{code-davinci-002} LLM \cite{chen2021evaluating} is used to parameterize a language model, with the capacity to assign conditional probabilities over any string $x_i\in\mathcal{X}$.
To interpret the score of each program $y_i\in\mathcal{Y}$ as a normalized probability with respect to the restricted hypothesis space under consideration, the log-probabilities of the considered programs under the LLM are passed through a softmax function with temperature parameter $\alpha$, selected independently for each stimulus sentence using leave-one-out cross-validation (LOOCV) as expanded in the following section.

\begin{equation}
P(y_i) = \frac{\exp \left(\alpha \log P(x_i) \right)}{\sum_{j=1}^n \exp \left( \alpha \log P(x_j) \right)}
\label{eqn:probs}
\end{equation}

In this case where programs differ only in $\theta$, $P_{model}(\theta_i)$ is approximated as $P(y_i)$.
These discrete program probabilities form the basis for subsequent analyses.

\subsection{Comparing $P_{human}(\theta)$ and $P_{model}(\theta)$}

For each of the 36 stimulus sentences, 29 (E1) or 30 (E2) point estimates on $\theta$ were measured in human participants.
From these point estimates, a discrete empirical distribution over the domain $0-100$, in intervals of $10$, was calculated via normalized counts for each stimulus.

\begin{equation}
P_{human}(\theta_i) = \frac{C(\theta_i)}{\sum_{j=1}^n C(\theta_j)}
\end{equation}

For the same stimulus sentences, a weight was calculated for each program over the same domain.
Such weights were normalized as in Equation \ref{eqn:probs} with $\alpha$ selected for each stimulus by minimizing the sum of the Jensen-Shannon distances (JSD; Equation \ref{eqn:jsd}) between $P_{human}(\theta)$ and $P_{model}(\theta)$ for the remaining $N-1$ stimuli per experiment, using the Nelder-Mead downhill simplex method \cite{nelder1965simplex}.

\begin{equation}
\argmin_{\alpha} JSD \left( P_{human}(\theta),P_{model}(\theta) \right)
\end{equation}

With $P_{human}(\theta)$ and $P_{model}(\theta)$ defined, their similarity was calculated using the Jensen-Shannon distance, a metric distance between two probability distributions $P$ and $Q$, where $M$ is the point-wise mean between $P$ and $Q$, and $KL$ is the Kullback-Leibler divergence \cite{lin1991divergence}.

\begin{equation}
JSD \left( P \parallel Q \right) = \sqrt{ \frac{ KL \left(P \parallel M \right) + KL \left( Q \parallel M \right)}{2} }
\label{eqn:jsd}
\end{equation}

In order to evaluate statistical significance of this similarity metric, a nonparametric permutation test was employed.
To generate the null distribution, the values of $P_{human}(\theta)$ and $P_{model}(\theta)$ were shuffled over $\theta$ for $N=10,000$ iterations and JSD measured for each variant.
$p$-values were calculated as the count of null samples less than the true JSD normalized by $N$.
Raw $p$-values were controlled for multiple comparisons using False discovery rate (FDR) correction for the number of tests, within each experiment \cite{benjamini1995controlling}.

\section{Results}

\begin{figure}[h]
\begin{center}
\includegraphics[width=0.49\textwidth]{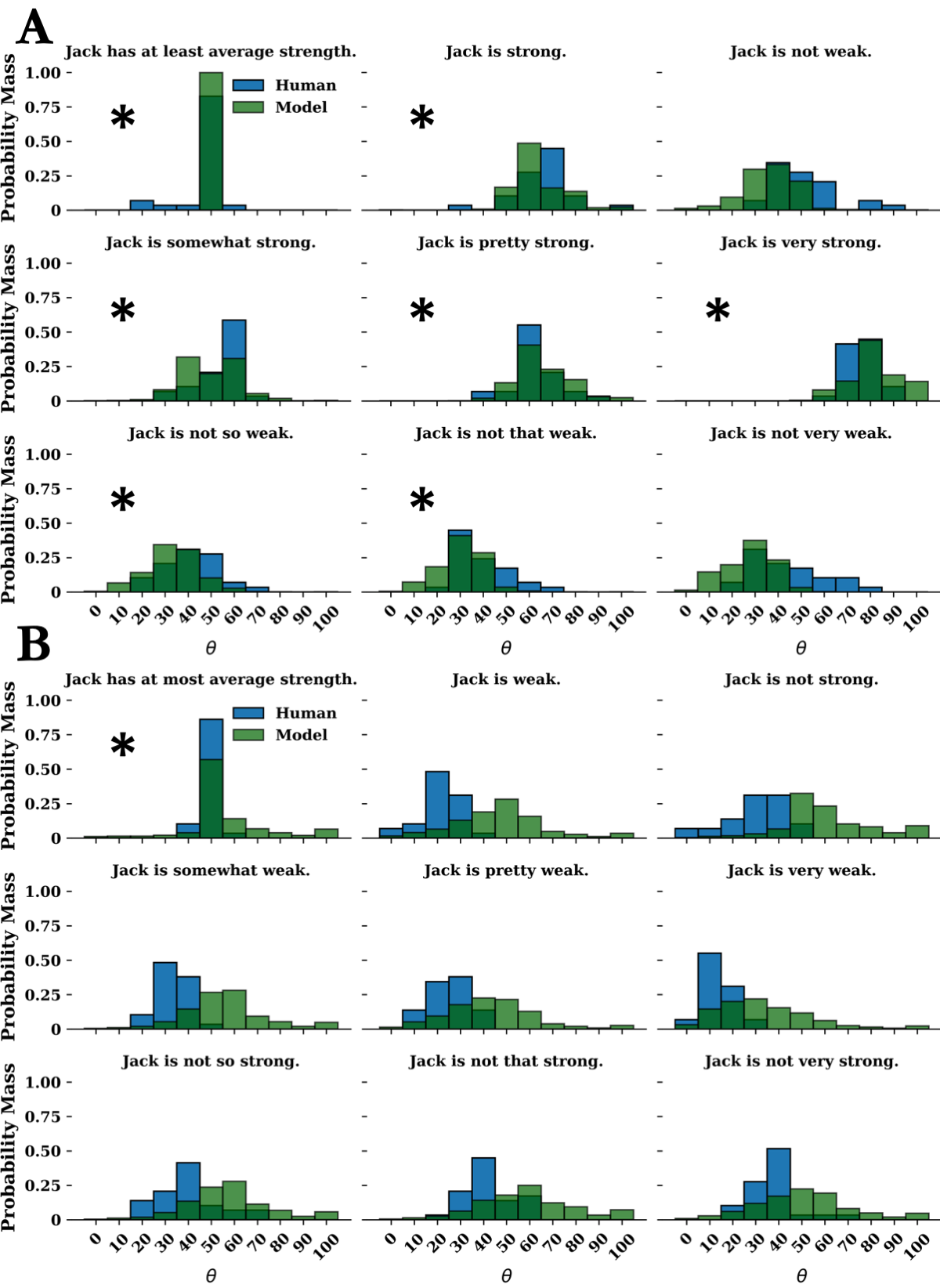}
\end{center}
\caption{Model-estimated and human-measured distributions over $P(\theta)$. Panel A explores programs of the form: \church{(condition (> (strength 'jack) $\theta$))}, and Panel B: \church{(condition (< (strength 'jack) $\theta$))}. Each subplot considers a unique sentence, with $P_{model}(\theta)$ presented in green and $P_{human}(\theta)$ in blue. An asterisk indicates significant similarity ($p<0.05$; FDR-corrected)  between $P_{model}(\theta)$ and $P_{human}(\theta)$, instantiated as a reduced Jensen-Shannon Distance (JSD; Equation \ref{eqn:jsd}) relative to a null permutation analysis. } 
\label{av1-plot}
\end{figure}

In order to evaluate whether LLMs can effectively leverage context to accurately infer distributions over linguistic meaning, several experiments were conducted.

\subsection{E1: Qualifiers}

For descriptions of Jack's strength, programs of the form \church{(condition (> (strength 'jack) $\theta$))} were evaluated over $\theta$.
$P_{model}(\theta)$ is presented in green in Figure \ref{av1-plot}A.

\subsubsection{LLMs make contextually-aware, pragmatically-sensitive inferences over graded adjectives and qualifiers.}
For each variation, a qualitatively smooth and interpretable distribution is reflected over $\theta$.
For \qtext{Jack is strong} the majority of probability mass falls $>\mu_{strength}$, and when \qtext{Jack is very strong} it shifts further.
For the control \qtext{Jack has at least average strength}, the mass is correctly placed on $\theta=\mu_{strength}$.

\subsubsection{LLMs make mostly human-like inferences, but struggle with negation.}
On the same Figure \ref{av1-plot}A, we see $P_{human}(\theta)$ presented in blue.
Remarkably, $P_{human}(\theta)$ and $P_{model}(\theta)$ are generally highly overlapping, even often with complex qualifier composition.
In fact, such distributions present with significant similarity for all sentences lacking negation (Figure \ref{av1-plot}A).
However, of the sentences including negation, only half of the interpretations are well-aligned.

\subsubsection{LLMs struggle further with polarity inversion.}
To further evaluate the robustness of this framework, a follow-up experiment was conducted, exploring inversion in concept polarity.
For a collection of sentences describing the Jack's weakness, programs of the form, \church{(condition (< (strength 'jack) $\theta$))} were evaluated over the domain of $\theta$.
$P_{model}(\theta)$ is presented in green in Figure \ref{av1-plot}B.
Once again, distributions appear qualitatively smooth and present with some intuitive characteristics.
For example, \qtext{Jack is very weak} is less than \qtext{Jack is weak}, and the mean is correctly parsed in the control \qtext{Jack has at most average strength.}
However, a different trend is observed with respect to the alignment with human participants.
In this case, where the evaluated concept is of negative polarity with respect to the variable presented in the prompt, $\theta$ tends to be consistently overestimated by the model.
For all sentences other than the control, there is an inability to detect significant similarity between $P_{model}(\theta)$ and $P_{human}(\theta)$.

\subsection{E2: Comparison Classes}

Selecting the control, \qtext{at least average}, and the condition deviated most from $\mu_{strength}$ in Figure \ref{av1-plot}A, \qtext{very strong}, a new set of sentences were compiled to describe the strength of \qtext{Jack} contingent on his membership in different \qtext{leagues} with individual strength priors.
The prompt explicitly presents \qtext{beginner}, \qtext{intermediate}, and \qtext{professional} leagues, with respective means of $30$, $50$, and $70$.

\subsubsection{LLMs accurately parse conditional mixtures, even inferring group membership from indirect descriptors.}
\begin{figure}[h]
\begin{center}
\includegraphics[width=0.49\textwidth]{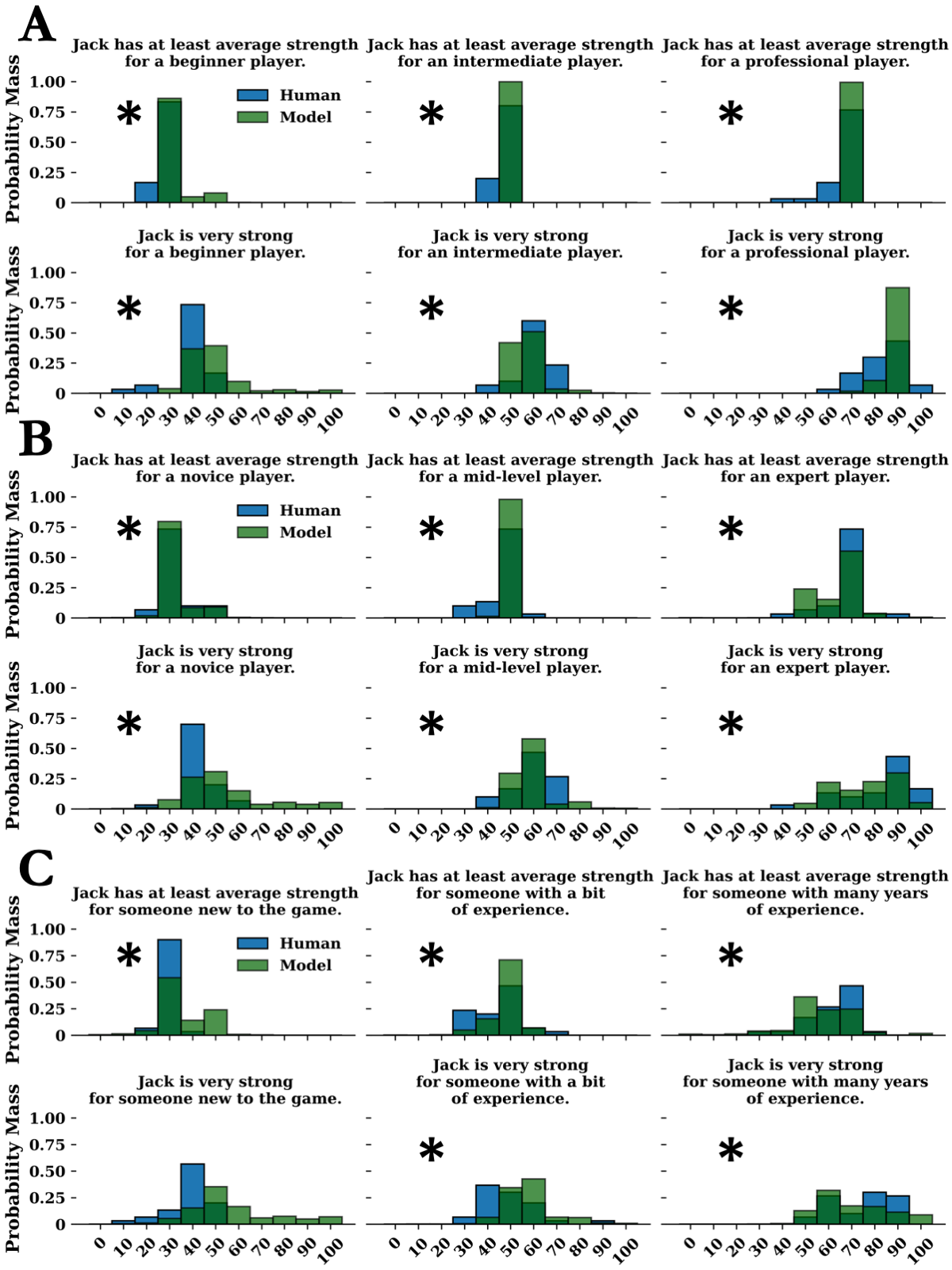}
\end{center}
\caption{Model-estimated and human-measured distributions over $P(\theta)$, incorporating comparison class. Panel A uses exact class from prompt, Panel B: synonyms, and Panel C: allusions. As in Figure \ref{av1-plot}, $P_{model}(\theta)$ is presented in green, $P_{human}(\theta)$ in blue, and an asterisk indicates significant ($p<0.05$; FDR) similarity between $P_{model}(\theta)$ and $P_{human}(\theta)$.} 
\label{av2-plot}
\end{figure}

Sentences of the form \qtext{Jack \dots for a \dots} were presented for each strength description and each league, including both the exact leagues described in the prompt (Figure \ref{av2-plot}A), as well as previously unseen league descriptors as synonyms (Figure \ref{av2-plot}B), and even indirect allusions (Figure \ref{av2-plot}C).
Such sentences were parsed and interpreted with outstanding success, significantly aligning with human expectations for 17 of the 18 sentences evaluated, including all control sentences and all sentences at the complexity of direct matches or synonyms.

\section{Discussion}

We began this work with a framework of pragmatic language understanding as an inferential procedure, and next motivated a view of linguistic meaning representation as probabilistic programs.
Selecting gradable adjectives as our test bed, we designed a task to evaluate the pragmatic reasoning capacity of LLMs in a complex semantic parsing exercise.
Contextualized on code expressing a generative world model defining the semantics of a tug-of-war game, we evaluated a number of sentences about the strength of a fictional player, often composing such sentences with pragmatically complex phenomena.
Using an LLM, we estimated $P_{model}(\theta)$ for each target sentence and conducted human behavioral experiments to empirically measure each corresponding $P_{human}(\theta)$.

From our initial evaluation (E1; Figure \ref{av1-plot}), we learned that LLMs can effectively amortize inference of a smooth distribution over $\theta$ in a way that is contextually-grounded to the semantics of the prompt and pragmatically-sensitive with respect to gradable adjectives and qualifiers.
Such model estimates aligned with human measurements for all descriptions of how \qtext{strong} a player was, but failed to recapitulate the intricacies of human distributions in the majority of cases where the player was \qtext{weak}, \qtext{not weak}, or \qtext{not strong}.
These results suggest that while the model can estimate \textit{some} approximate distribution for each of these cases, the ability to infer an exactly human-like distribution suffers when composing negation in the lexical space, e.g., \qtext{strong} vs. \qtext{not strong}, or polarity inversion in the conceptual space, e.g., \qtext{strong} vs. \qtext{weak}.
This is consistent with prior work noting LLM difficulty in resolving negation more generally \cite{kassner2019negated,hosseini2021understanding,creswell2022selection}.
It also draws intriguing parallels to child developmental work on concept acquisition, noting observed lags in the mastery of negative polarity concepts, e.g., \qtext{short}, relative to their positive polarity counterparts, e.g., \qtext{tall}, perhaps highlighting more general asymmetries in concept complexity \cite{klatzky1973asymmetries,brewer1975acquisition,barner2008compositionality}.
From our evaluation of class comparisons (E2; Figure \ref{av2-plot}), we further highlighted the context-sensitivity of such models in appropriately resolving conditional mixtures, presenting with impressive robustness in the presence of incorrect references nearby in context.
These results are even more powerful when the match between the query and context variable is not exact, but instead needs to be estimated from a synonym or indirect allusion.
These results support an argument for the lexical semantic robustness of LLMs under this approach, a convenient case relative to some traditional semantic parsers based on combinatory categorical grammars (CCGs), for which more complex workarounds are often required \cite{artzi2014learning,kwiatkowski2011lexical,steedman2001syntactic}.

Overall, these results paint a picture of LLMs as effectively recovering some reasonable distribution in each of these complex test cases, yet highlight some discrepancies with human inferences.
If we had perfectly recovered human distributions, this would have led to a series of possible interpretations.
One interpretation of such a finding might be that LLMs, just as they appear to implicitly represent other forms of linguistic structure, here implicitly perform inference, as alluded to via other works on amortization \cite{white2020learning,wu2022foundation}.
Another interpretation could be that, in practice, the statistical regularities of text during training are sufficient to recover these distributions at test time without explicit computation over a world model.
Such an account might inform resource-rational frameworks of human language processing, possibly suggesting that partial pragmatic computations could in principle be heuristically approximated, or even retrieved, instead of explicitly recomputed at each instance \cite{gershman2014amortized,gershman2015computational,dasgupta2021memory}.
While our data do not present LLMs as perfect estimates of human populations across all cases, we believe that these data still at least partially support this second hypothesis.
It is indeed possible that some, but not all, of the computations required to solve our task, are amortizable, lending to human-like distributions in some cases, but incorrect approximation in other out-of-domain cases.
For example, perhaps composition with negation requires more explicit computation at test time by human participants, which leads to this distributional shift relative to the heuristic estimate of the LLMs.
Future work should consider more directly testing this, starting from a framework of computational utility.

\subsubsection{Limitations}

While the results presented in this work have proposed a primarily positive image of LLMs as elegantly handling pragmatic inference within a complex semantic parsing task, only a small number of examples within a single scope have been explored thus far.
In order to confirm that the conclusions of these results generalize, evaluation of a broader class of pragmatic phenomena in additional task contexts would be required.

\subsubsection{Future Directions}

One particularly exciting future direction is connecting LLM-mediated inferences over PPL programs with actual execution of such programs and evaluation of their resulting distributions.
If we ask \qtext{Can Jill, a very strong beginner, beat Jane, a somewhat strong intermediate?}, such a question can be reduced to neuro-symbolic programming.
Leveraging an LLM inference, a distribution over the thresholds on each players' strengths can be derived.
Next, such programs can be explicitly executed in a PPL interpreter, inducing a distribution over each player strength.
From this state, it follows easily to query the winner of such a match: \church{(query (won-against '(jill) '(jane)))}.
In then considering more difficult cases, e.g., those involving negation, a hybrid between RSA-like and LLM-mediated approaches might be considered.
For example, using LLM estimates to initialize Sequential Monte Carlo (SMC) hypotheses that get updated based on probabilistic program inferences.

\newpage

\section{Acknowledgments}

We thank our anonymous reviewers for their insightful feedback and recommendations.
BL is supported by an MIT Presidential Fellowship and GG by the National Science Foundation Graduate Research Fellowship under Grant No. 2141064.
LW and JBT are supported by the MIT Quest for Intelligence, AFOSR Grant \#FA9550-19-1-0269, the MIT-IBM Watson AI Lab, ONR Science of AI and DARPA Machine Common Sense.

\bibliographystyle{apacite}
\setlength{\bibleftmargin}{.125in}
\setlength{\bibindent}{-\bibleftmargin}
\bibliography{main}

\begin{thebibliography}{}

\bibitem [\protect \citeauthoryear {%
Acquaviva%
\ \protect \BOthers {.}}{%
Acquaviva%
\ \protect \BOthers {.}}{%
{\protect \APACyear {2021}}%
}]{%
acquaviva2021communicating}
\APACinsertmetastar {%
acquaviva2021communicating}%
\begin{APACrefauthors}%
Acquaviva, S.%
, Pu, Y.%
, Kryven, M.%
, Sechopoulos, T.%
, Wong, C.%
, Ecanow, G\BPBI E.%
\BDBL {}Tenenbaum, J\BPBI B.%
\end{APACrefauthors}%
\unskip\
\newblock
\APACrefYearMonthDay{2021}{}{}.
\newblock
{\BBOQ}\APACrefatitle {Communicating natural programs to humans and machines}
  {Communicating natural programs to humans and machines}.{\BBCQ}
\newblock
\APACjournalVolNumPages{arXiv preprint arXiv:2106.07824}{}{}{}.
\PrintBackRefs{\CurrentBib}

\bibitem [\protect \citeauthoryear {%
Artzi%
, Das%
\BCBL {}\ \BBA {} Petrov%
}{%
Artzi%
\ \protect \BOthers {.}}{%
{\protect \APACyear {2014}}%
}]{%
artzi2014learning}
\APACinsertmetastar {%
artzi2014learning}%
\begin{APACrefauthors}%
Artzi, Y.%
, Das, D.%
\BCBL {}\ \BBA {} Petrov, S.%
\end{APACrefauthors}%
\unskip\
\newblock
\APACrefYearMonthDay{2014}{}{}.
\newblock
{\BBOQ}\APACrefatitle {Learning compact lexicons for CCG semantic parsing}
  {Learning compact lexicons for ccg semantic parsing}.{\BBCQ}
\newblock

\PrintBackRefs{\CurrentBib}

\bibitem [\protect \citeauthoryear {%
Austin%
}{%
Austin%
}{%
{\protect \APACyear {1975}}%
}]{%
austin1975things}
\APACinsertmetastar {%
austin1975things}%
\begin{APACrefauthors}%
Austin, J\BPBI L.%
\end{APACrefauthors}%
\unskip\
\newblock
\APACrefYear{1975}.
\newblock
\APACrefbtitle {How to do things with words} {How to do things with words}.
\newblock
\APACaddressPublisher{}{Oxford university press}.
\PrintBackRefs{\CurrentBib}

\bibitem [\protect \citeauthoryear {%
Barner%
\ \BBA {} Snedeker%
}{%
Barner%
\ \BBA {} Snedeker%
}{%
{\protect \APACyear {2008}}%
}]{%
barner2008compositionality}
\APACinsertmetastar {%
barner2008compositionality}%
\begin{APACrefauthors}%
Barner, D.%
\BCBT {}\ \BBA {} Snedeker, J.%
\end{APACrefauthors}%
\unskip\
\newblock
\APACrefYearMonthDay{2008}{}{}.
\newblock
{\BBOQ}\APACrefatitle {Compositionality and statistics in adjective
  acquisition: 4-year-olds interpret tall and short based on the size
  distributions of novel noun referents} {Compositionality and statistics in
  adjective acquisition: 4-year-olds interpret tall and short based on the size
  distributions of novel noun referents}.{\BBCQ}
\newblock
\APACjournalVolNumPages{Child development}{79}{3}{594--608}.
\PrintBackRefs{\CurrentBib}

\bibitem [\protect \citeauthoryear {%
Benjamini%
\ \BBA {} Hochberg%
}{%
Benjamini%
\ \BBA {} Hochberg%
}{%
{\protect \APACyear {1995}}%
}]{%
benjamini1995controlling}
\APACinsertmetastar {%
benjamini1995controlling}%
\begin{APACrefauthors}%
Benjamini, Y.%
\BCBT {}\ \BBA {} Hochberg, Y.%
\end{APACrefauthors}%
\unskip\
\newblock
\APACrefYearMonthDay{1995}{}{}.
\newblock
{\BBOQ}\APACrefatitle {Controlling the false discovery rate: a practical and
  powerful approach to multiple testing} {Controlling the false discovery rate:
  a practical and powerful approach to multiple testing}.{\BBCQ}
\newblock
\APACjournalVolNumPages{Journal of the Royal statistical society: series B
  (Methodological)}{57}{1}{289--300}.
\PrintBackRefs{\CurrentBib}

\bibitem [\protect \citeauthoryear {%
Bommasani%
\ \protect \BOthers {.}}{%
Bommasani%
\ \protect \BOthers {.}}{%
{\protect \APACyear {2021}}%
}]{%
bommasani2021opportunities}
\APACinsertmetastar {%
bommasani2021opportunities}%
\begin{APACrefauthors}%
Bommasani, R.%
, Hudson, D\BPBI A.%
, Adeli, E.%
, Altman, R.%
, Arora, S.%
, von Arx, S.%
\BDBL {}others%
\end{APACrefauthors}%
\unskip\
\newblock
\APACrefYearMonthDay{2021}{}{}.
\newblock
{\BBOQ}\APACrefatitle {On the opportunities and risks of foundation models} {On
  the opportunities and risks of foundation models}.{\BBCQ}
\newblock
\APACjournalVolNumPages{arXiv preprint arXiv:2108.07258}{}{}{}.
\PrintBackRefs{\CurrentBib}

\bibitem [\protect \citeauthoryear {%
Brewer%
\ \BBA {} Stone%
}{%
Brewer%
\ \BBA {} Stone%
}{%
{\protect \APACyear {1975}}%
}]{%
brewer1975acquisition}
\APACinsertmetastar {%
brewer1975acquisition}%
\begin{APACrefauthors}%
Brewer, W\BPBI F.%
\BCBT {}\ \BBA {} Stone, J\BPBI B.%
\end{APACrefauthors}%
\unskip\
\newblock
\APACrefYearMonthDay{1975}{}{}.
\newblock
{\BBOQ}\APACrefatitle {Acquisition of spatial antonym pairs} {Acquisition of
  spatial antonym pairs}.{\BBCQ}
\newblock
\APACjournalVolNumPages{Journal of Experimental Child
  Psychology}{19}{2}{299--307}.
\PrintBackRefs{\CurrentBib}

\bibitem [\protect \citeauthoryear {%
Chen%
\ \protect \BOthers {.}}{%
Chen%
\ \protect \BOthers {.}}{%
{\protect \APACyear {2021}}%
}]{%
chen2021evaluating}
\APACinsertmetastar {%
chen2021evaluating}%
\begin{APACrefauthors}%
Chen, M.%
, Tworek, J.%
, Jun, H.%
, Yuan, Q.%
, Pinto, H\BPBI P\BPBI d\BPBI O.%
, Kaplan, J.%
\BDBL {}others%
\end{APACrefauthors}%
\unskip\
\newblock
\APACrefYearMonthDay{2021}{}{}.
\newblock
{\BBOQ}\APACrefatitle {Evaluating large language models trained on code}
  {Evaluating large language models trained on code}.{\BBCQ}
\newblock
\APACjournalVolNumPages{arXiv}{}{}{}.
\PrintBackRefs{\CurrentBib}

\bibitem [\protect \citeauthoryear {%
Clark%
}{%
Clark%
}{%
{\protect \APACyear {1996}}%
}]{%
clark1996using}
\APACinsertmetastar {%
clark1996using}%
\begin{APACrefauthors}%
Clark, H\BPBI H.%
\end{APACrefauthors}%
\unskip\
\newblock
\APACrefYear{1996}.
\newblock
\APACrefbtitle {Using language} {Using language}.
\newblock
\APACaddressPublisher{}{Cambridge university press}.
\PrintBackRefs{\CurrentBib}

\bibitem [\protect \citeauthoryear {%
Collins%
, Wong%
, Feng%
, Wei%
\BCBL {}\ \BBA {} Tenenbaum%
}{%
Collins%
\ \protect \BOthers {.}}{%
{\protect \APACyear {2022}}%
}]{%
collins2022structured}
\APACinsertmetastar {%
collins2022structured}%
\begin{APACrefauthors}%
Collins, K\BPBI M.%
, Wong, C.%
, Feng, J.%
, Wei, M.%
\BCBL {}\ \BBA {} Tenenbaum, J\BPBI B.%
\end{APACrefauthors}%
\unskip\
\newblock
\APACrefYearMonthDay{2022}{}{}.
\newblock
{\BBOQ}\APACrefatitle {Structured, flexible, and robust: benchmarking and
  improving large language models towards more human-like behavior in
  out-of-distribution reasoning tasks} {Structured, flexible, and robust:
  benchmarking and improving large language models towards more human-like
  behavior in out-of-distribution reasoning tasks}.{\BBCQ}
\newblock
\APACjournalVolNumPages{arXiv preprint arXiv:2205.05718}{}{}{}.
\PrintBackRefs{\CurrentBib}

\bibitem [\protect \citeauthoryear {%
Cooper%
, Dobnik%
, Lappin%
\BCBL {}\ \BBA {} Larsson%
}{%
Cooper%
\ \protect \BOthers {.}}{%
{\protect \APACyear {2015}}%
}]{%
cooper2015probabilistic}
\APACinsertmetastar {%
cooper2015probabilistic}%
\begin{APACrefauthors}%
Cooper, R.%
, Dobnik, S.%
, Lappin, S.%
\BCBL {}\ \BBA {} Larsson, S.%
\end{APACrefauthors}%
\unskip\
\newblock
\APACrefYearMonthDay{2015}{}{}.
\newblock
{\BBOQ}\APACrefatitle {Probabilistic type theory and natural language
  semantics} {Probabilistic type theory and natural language semantics}.{\BBCQ}
\newblock
\APACjournalVolNumPages{Linguistic issues in language technology}{10}{}{}.
\PrintBackRefs{\CurrentBib}

\bibitem [\protect \citeauthoryear {%
Cresswell%
}{%
Cresswell%
}{%
{\protect \APACyear {1976}}%
}]{%
cresswell1976semantics}
\APACinsertmetastar {%
cresswell1976semantics}%
\begin{APACrefauthors}%
Cresswell, M\BPBI J.%
\end{APACrefauthors}%
\unskip\
\newblock
\APACrefYearMonthDay{1976}{}{}.
\newblock
{\BBOQ}\APACrefatitle {The semantics of degree} {The semantics of
  degree}.{\BBCQ}
\newblock
\BIn{} \APACrefbtitle {Montague grammar} {Montague grammar}\ (\BPGS\ 261--292).
\newblock
\APACaddressPublisher{}{Elsevier}.
\PrintBackRefs{\CurrentBib}

\bibitem [\protect \citeauthoryear {%
Creswell%
, Shanahan%
\BCBL {}\ \BBA {} Higgins%
}{%
Creswell%
\ \protect \BOthers {.}}{%
{\protect \APACyear {2022}}%
}]{%
creswell2022selection}
\APACinsertmetastar {%
creswell2022selection}%
\begin{APACrefauthors}%
Creswell, A.%
, Shanahan, M.%
\BCBL {}\ \BBA {} Higgins, I.%
\end{APACrefauthors}%
\unskip\
\newblock
\APACrefYearMonthDay{2022}{}{}.
\newblock
{\BBOQ}\APACrefatitle {Selection-inference: Exploiting large language models
  for interpretable logical reasoning} {Selection-inference: Exploiting large
  language models for interpretable logical reasoning}.{\BBCQ}
\newblock
\APACjournalVolNumPages{arXiv preprint arXiv:2205.09712}{}{}{}.
\PrintBackRefs{\CurrentBib}

\bibitem [\protect \citeauthoryear {%
Dasgupta%
\ \BBA {} Gershman%
}{%
Dasgupta%
\ \BBA {} Gershman%
}{%
{\protect \APACyear {2021}}%
}]{%
dasgupta2021memory}
\APACinsertmetastar {%
dasgupta2021memory}%
\begin{APACrefauthors}%
Dasgupta, I.%
\BCBT {}\ \BBA {} Gershman, S\BPBI J.%
\end{APACrefauthors}%
\unskip\
\newblock
\APACrefYearMonthDay{2021}{}{}.
\newblock
{\BBOQ}\APACrefatitle {Memory as a computational resource} {Memory as a
  computational resource}.{\BBCQ}
\newblock
\APACjournalVolNumPages{Trends in Cognitive Sciences}{25}{3}{240--251}.
\PrintBackRefs{\CurrentBib}

\bibitem [\protect \citeauthoryear {%
Frank%
\ \BBA {} Goodman%
}{%
Frank%
\ \BBA {} Goodman%
}{%
{\protect \APACyear {2012}}%
}]{%
frank2012predicting}
\APACinsertmetastar {%
frank2012predicting}%
\begin{APACrefauthors}%
Frank, M.%
\BCBT {}\ \BBA {} Goodman, N\BPBI D.%
\end{APACrefauthors}%
\unskip\
\newblock
\APACrefYearMonthDay{2012}{}{}.
\newblock
{\BBOQ}\APACrefatitle {Predicting pragmatic reasoning in language games}
  {Predicting pragmatic reasoning in language games}.{\BBCQ}
\newblock
\APACjournalVolNumPages{Science}{336}{6084}{998--998}.
\PrintBackRefs{\CurrentBib}

\bibitem [\protect \citeauthoryear {%
Fried%
, Tomlin%
, Hu%
, Patel%
\BCBL {}\ \BBA {} Nematzadeh%
}{%
Fried%
\ \protect \BOthers {.}}{%
{\protect \APACyear {2022}}%
}]{%
fried2022pragmatics}
\APACinsertmetastar {%
fried2022pragmatics}%
\begin{APACrefauthors}%
Fried, D.%
, Tomlin, N.%
, Hu, J.%
, Patel, R.%
\BCBL {}\ \BBA {} Nematzadeh, A.%
\end{APACrefauthors}%
\unskip\
\newblock
\APACrefYearMonthDay{2022}{}{}.
\newblock
{\BBOQ}\APACrefatitle {Pragmatics in Grounded Language Learning: Phenomena,
  Tasks, and Modeling Approaches} {Pragmatics in grounded language learning:
  Phenomena, tasks, and modeling approaches}.{\BBCQ}
\newblock
\APACjournalVolNumPages{arXiv preprint arXiv:2211.08371}{}{}{}.
\PrintBackRefs{\CurrentBib}

\bibitem [\protect \citeauthoryear {%
Gao%
\ \protect \BOthers {.}}{%
Gao%
\ \protect \BOthers {.}}{%
{\protect \APACyear {2022}}%
}]{%
gao2022pal}
\APACinsertmetastar {%
gao2022pal}%
\begin{APACrefauthors}%
Gao, L.%
, Madaan, A.%
, Zhou, S.%
, Alon, U.%
, Liu, P.%
, Yang, Y.%
\BDBL {}Neubig, G.%
\end{APACrefauthors}%
\unskip\
\newblock
\APACrefYearMonthDay{2022}{}{}.
\newblock
{\BBOQ}\APACrefatitle {PAL: Program-aided Language Models} {Pal: Program-aided
  language models}.{\BBCQ}
\newblock
\APACjournalVolNumPages{arXiv}{}{}{}.
\PrintBackRefs{\CurrentBib}

\bibitem [\protect \citeauthoryear {%
Gershman%
\ \BBA {} Goodman%
}{%
Gershman%
\ \BBA {} Goodman%
}{%
{\protect \APACyear {2014}}%
}]{%
gershman2014amortized}
\APACinsertmetastar {%
gershman2014amortized}%
\begin{APACrefauthors}%
Gershman, S.%
\BCBT {}\ \BBA {} Goodman, N.%
\end{APACrefauthors}%
\unskip\
\newblock
\APACrefYearMonthDay{2014}{}{}.
\newblock
{\BBOQ}\APACrefatitle {Amortized inference in probabilistic reasoning}
  {Amortized inference in probabilistic reasoning}.{\BBCQ}
\newblock
\BIn{} \APACrefbtitle {Proceedings of the annual meeting of the cognitive
  science society} {Proceedings of the annual meeting of the cognitive science
  society}\ (\BVOL~36).
\PrintBackRefs{\CurrentBib}

\bibitem [\protect \citeauthoryear {%
Gershman%
, Horvitz%
\BCBL {}\ \BBA {} Tenenbaum%
}{%
Gershman%
\ \protect \BOthers {.}}{%
{\protect \APACyear {2015}}%
}]{%
gershman2015computational}
\APACinsertmetastar {%
gershman2015computational}%
\begin{APACrefauthors}%
Gershman, S.%
, Horvitz, E\BPBI J.%
\BCBL {}\ \BBA {} Tenenbaum, J\BPBI B.%
\end{APACrefauthors}%
\unskip\
\newblock
\APACrefYearMonthDay{2015}{}{}.
\newblock
{\BBOQ}\APACrefatitle {Computational rationality: A converging paradigm for
  intelligence in brains, minds, and machines} {Computational rationality: A
  converging paradigm for intelligence in brains, minds, and machines}.{\BBCQ}
\newblock
\APACjournalVolNumPages{Science}{349}{6245}{273--278}.
\PrintBackRefs{\CurrentBib}

\bibitem [\protect \citeauthoryear {%
Gerstenberg%
\ \BBA {} Goodman%
}{%
Gerstenberg%
\ \BBA {} Goodman%
}{%
{\protect \APACyear {2012}}%
}]{%
gerstenberg2012ping}
\APACinsertmetastar {%
gerstenberg2012ping}%
\begin{APACrefauthors}%
Gerstenberg, T.%
\BCBT {}\ \BBA {} Goodman, N\BPBI D.%
\end{APACrefauthors}%
\unskip\
\newblock
\APACrefYearMonthDay{2012}{}{}.
\newblock
{\BBOQ}\APACrefatitle {Ping Pong in Church: Productive use of concepts in human
  probabilistic inference} {Ping pong in church: Productive use of concepts in
  human probabilistic inference}.{\BBCQ}
\newblock
\BIn{} \APACrefbtitle {Proceedings of the annual meeting of the cognitive
  science society} {Proceedings of the annual meeting of the cognitive science
  society}\ (\BVOL~34).
\PrintBackRefs{\CurrentBib}

\bibitem [\protect \citeauthoryear {%
Goodman%
\ \BBA {} Frank%
}{%
Goodman%
\ \BBA {} Frank%
}{%
{\protect \APACyear {2016}}%
}]{%
goodman2016pragmatic}
\APACinsertmetastar {%
goodman2016pragmatic}%
\begin{APACrefauthors}%
Goodman, N\BPBI D.%
\BCBT {}\ \BBA {} Frank, M\BPBI C.%
\end{APACrefauthors}%
\unskip\
\newblock
\APACrefYearMonthDay{2016}{}{}.
\newblock
{\BBOQ}\APACrefatitle {Pragmatic language interpretation as probabilistic
  inference} {Pragmatic language interpretation as probabilistic
  inference}.{\BBCQ}
\newblock
\APACjournalVolNumPages{Trends in cognitive sciences}{20}{11}{818--829}.
\PrintBackRefs{\CurrentBib}

\bibitem [\protect \citeauthoryear {%
Goodman%
\ \BBA {} Lassiter%
}{%
Goodman%
\ \BBA {} Lassiter%
}{%
{\protect \APACyear {2015}}%
}]{%
goodman2015probabilistic}
\APACinsertmetastar {%
goodman2015probabilistic}%
\begin{APACrefauthors}%
Goodman, N\BPBI D.%
\BCBT {}\ \BBA {} Lassiter, D.%
\end{APACrefauthors}%
\unskip\
\newblock
\APACrefYearMonthDay{2015}{}{}.
\newblock
{\BBOQ}\APACrefatitle {Probabilistic semantics and pragmatics: Uncertainty in
  language and thought} {Probabilistic semantics and pragmatics: Uncertainty in
  language and thought}.{\BBCQ}
\newblock
\APACjournalVolNumPages{The handbook of contemporary semantic theory, 2nd
  edition. Wiley-Blackwell}{}{}{}.
\PrintBackRefs{\CurrentBib}

\bibitem [\protect \citeauthoryear {%
Goodman%
, Mansinghka%
, Roy%
, Bonawitz%
\BCBL {}\ \BBA {} Tenenbaum%
}{%
Goodman%
\ \protect \BOthers {.}}{%
{\protect \APACyear {2012}}%
}]{%
goodman2012church}
\APACinsertmetastar {%
goodman2012church}%
\begin{APACrefauthors}%
Goodman, N\BPBI D.%
, Mansinghka, V.%
, Roy, D.%
, Bonawitz, K.%
\BCBL {}\ \BBA {} Tenenbaum, J\BPBI B.%
\end{APACrefauthors}%
\unskip\
\newblock
\APACrefYearMonthDay{2012}{}{}.
\newblock
{\BBOQ}\APACrefatitle {Church: a language for generative models} {Church: a
  language for generative models}.{\BBCQ}
\newblock
\APACjournalVolNumPages{arXiv}{}{}{}.
\PrintBackRefs{\CurrentBib}

\bibitem [\protect \citeauthoryear {%
Goodman%
\ \BBA {} Stuhlm{\"u}ller%
}{%
Goodman%
\ \BBA {} Stuhlm{\"u}ller%
}{%
{\protect \APACyear {2013}}%
}]{%
goodman2013knowledge}
\APACinsertmetastar {%
goodman2013knowledge}%
\begin{APACrefauthors}%
Goodman, N\BPBI D.%
\BCBT {}\ \BBA {} Stuhlm{\"u}ller, A.%
\end{APACrefauthors}%
\unskip\
\newblock
\APACrefYearMonthDay{2013}{}{}.
\newblock
{\BBOQ}\APACrefatitle {Knowledge and implicature: Modeling language
  understanding as social cognition} {Knowledge and implicature: Modeling
  language understanding as social cognition}.{\BBCQ}
\newblock
\APACjournalVolNumPages{Topics in cognitive science}{5}{1}{173--184}.
\PrintBackRefs{\CurrentBib}

\bibitem [\protect \citeauthoryear {%
Goodman%
\ \BBA {} Tenenbaum%
}{%
Goodman%
\ \BBA {} Tenenbaum%
}{%
{\protect \APACyear {2010}}%
}]{%
probmods}
\APACinsertmetastar {%
probmods}%
\begin{APACrefauthors}%
Goodman, N\BPBI D.%
\BCBT {}\ \BBA {} Tenenbaum, J\BPBI B.%
\end{APACrefauthors}%
\unskip\
\newblock
\APACrefYearMonthDay{2010}{}{}.
\newblock
\APACrefbtitle {{Probabilistic Models of Cognition}} {{Probabilistic Models of
  Cognition}}\ (\PrintOrdinal{First}\ \BEd).
\newblock
\APAChowpublished {\url{http://v1.probmods.org/}}.
\PrintBackRefs{\CurrentBib}

\bibitem [\protect \citeauthoryear {%
Goodman%
, Tenenbaum%
\BCBL {}\ \BBA {} Gerstenberg%
}{%
Goodman%
\ \protect \BOthers {.}}{%
{\protect \APACyear {2014}}%
}]{%
goodman2014concepts}
\APACinsertmetastar {%
goodman2014concepts}%
\begin{APACrefauthors}%
Goodman, N\BPBI D.%
, Tenenbaum, J\BPBI B.%
\BCBL {}\ \BBA {} Gerstenberg, T.%
\end{APACrefauthors}%
\unskip\
\newblock
\APACrefYearMonthDay{2014}{}{}.
\newblock
{\BBOQ}\APACrefatitle {Concepts in a probabilistic language of thought}
  {Concepts in a probabilistic language of thought}.{\BBCQ}
\newblock
\APACjournalVolNumPages{Center for Brains, Minds and Machines (CBMM)
  Memos}{010}{}{}.
\PrintBackRefs{\CurrentBib}

\bibitem [\protect \citeauthoryear {%
Grice%
}{%
Grice%
}{%
{\protect \APACyear {1989}}%
}]{%
grice1989studies}
\APACinsertmetastar {%
grice1989studies}%
\begin{APACrefauthors}%
Grice, P.%
\end{APACrefauthors}%
\unskip\
\newblock
\APACrefYear{1989}.
\newblock
\APACrefbtitle {Studies in the Way of Words} {Studies in the way of words}.
\newblock
\APACaddressPublisher{}{Harvard University Press}.
\PrintBackRefs{\CurrentBib}

\bibitem [\protect \citeauthoryear {%
Hosseini%
\ \protect \BOthers {.}}{%
Hosseini%
\ \protect \BOthers {.}}{%
{\protect \APACyear {2021}}%
}]{%
hosseini2021understanding}
\APACinsertmetastar {%
hosseini2021understanding}%
\begin{APACrefauthors}%
Hosseini, A.%
, Reddy, S.%
, Bahdanau, D.%
, Hjelm, R\BPBI D.%
, Sordoni, A.%
\BCBL {}\ \BBA {} Courville, A.%
\end{APACrefauthors}%
\unskip\
\newblock
\APACrefYearMonthDay{2021}{}{}.
\newblock
{\BBOQ}\APACrefatitle {Understanding by understanding not: Modeling negation in
  language models} {Understanding by understanding not: Modeling negation in
  language models}.{\BBCQ}
\newblock
\APACjournalVolNumPages{arXiv preprint arXiv:2105.03519}{}{}{}.
\PrintBackRefs{\CurrentBib}

\bibitem [\protect \citeauthoryear {%
Hu%
, Floyd%
, Jouravlev%
, Fedorenko%
\BCBL {}\ \BBA {} Gibson%
}{%
Hu%
\ \protect \BOthers {.}}{%
{\protect \APACyear {2022}}%
}]{%
hu2022fine}
\APACinsertmetastar {%
hu2022fine}%
\begin{APACrefauthors}%
Hu, J.%
, Floyd, S.%
, Jouravlev, O.%
, Fedorenko, E.%
\BCBL {}\ \BBA {} Gibson, E.%
\end{APACrefauthors}%
\unskip\
\newblock
\APACrefYearMonthDay{2022}{}{}.
\newblock
{\BBOQ}\APACrefatitle {A fine-grained comparison of pragmatic language
  understanding in humans and language models} {A fine-grained comparison of
  pragmatic language understanding in humans and language models}.{\BBCQ}
\newblock
\APACjournalVolNumPages{arXiv preprint arXiv:2212.06801}{}{}{}.
\PrintBackRefs{\CurrentBib}

\bibitem [\protect \citeauthoryear {%
Hu%
, Levy%
, Degen%
\BCBL {}\ \BBA {} Schuster%
}{%
Hu%
\ \protect \BOthers {.}}{%
{\protect \APACyear {2023}}%
}]{%
hu2023expectations}
\APACinsertmetastar {%
hu2023expectations}%
\begin{APACrefauthors}%
Hu, J.%
, Levy, R.%
, Degen, J.%
\BCBL {}\ \BBA {} Schuster, S.%
\end{APACrefauthors}%
\unskip\
\newblock
\APACrefYearMonthDay{2023}{}{}.
\newblock
{\BBOQ}\APACrefatitle {Expectations over Unspoken Alternatives Predict
  Pragmatic Inferences} {Expectations over unspoken alternatives predict
  pragmatic inferences}.{\BBCQ}
\newblock
\APACjournalVolNumPages{arXiv preprint arXiv:2304.04758}{}{}{}.
\PrintBackRefs{\CurrentBib}

\bibitem [\protect \citeauthoryear {%
Kassner%
\ \BBA {} Sch{\"u}tze%
}{%
Kassner%
\ \BBA {} Sch{\"u}tze%
}{%
{\protect \APACyear {2019}}%
}]{%
kassner2019negated}
\APACinsertmetastar {%
kassner2019negated}%
\begin{APACrefauthors}%
Kassner, N.%
\BCBT {}\ \BBA {} Sch{\"u}tze, H.%
\end{APACrefauthors}%
\unskip\
\newblock
\APACrefYearMonthDay{2019}{}{}.
\newblock
{\BBOQ}\APACrefatitle {Negated and misprimed probes for pretrained language
  models: Birds can talk, but cannot fly} {Negated and misprimed probes for
  pretrained language models: Birds can talk, but cannot fly}.{\BBCQ}
\newblock
\APACjournalVolNumPages{arXiv preprint arXiv:1911.03343}{}{}{}.
\PrintBackRefs{\CurrentBib}

\bibitem [\protect \citeauthoryear {%
Kennedy%
}{%
Kennedy%
}{%
{\protect \APACyear {2007}}%
}]{%
kennedy2007vagueness}
\APACinsertmetastar {%
kennedy2007vagueness}%
\begin{APACrefauthors}%
Kennedy, C.%
\end{APACrefauthors}%
\unskip\
\newblock
\APACrefYearMonthDay{2007}{}{}.
\newblock
{\BBOQ}\APACrefatitle {Vagueness and grammar: The semantics of relative and
  absolute gradable adjectives} {Vagueness and grammar: The semantics of
  relative and absolute gradable adjectives}.{\BBCQ}
\newblock
\APACjournalVolNumPages{Linguistics and philosophy}{30}{1}{1--45}.
\PrintBackRefs{\CurrentBib}

\bibitem [\protect \citeauthoryear {%
Klatzky%
, Clark%
\BCBL {}\ \BBA {} Macken%
}{%
Klatzky%
\ \protect \BOthers {.}}{%
{\protect \APACyear {1973}}%
}]{%
klatzky1973asymmetries}
\APACinsertmetastar {%
klatzky1973asymmetries}%
\begin{APACrefauthors}%
Klatzky, R\BPBI L.%
, Clark, E\BPBI V.%
\BCBL {}\ \BBA {} Macken, M.%
\end{APACrefauthors}%
\unskip\
\newblock
\APACrefYearMonthDay{1973}{}{}.
\newblock
{\BBOQ}\APACrefatitle {Asymmetries in the acquisition of polar adjectives:
  linguistic or conceptual?} {Asymmetries in the acquisition of polar
  adjectives: linguistic or conceptual?}{\BBCQ}
\newblock
\APACjournalVolNumPages{Journal of Experimental Child
  Psychology}{16}{1}{32--46}.
\PrintBackRefs{\CurrentBib}

\bibitem [\protect \citeauthoryear {%
Klein%
}{%
Klein%
}{%
{\protect \APACyear {1980}}%
}]{%
klein1980semantics}
\APACinsertmetastar {%
klein1980semantics}%
\begin{APACrefauthors}%
Klein, E.%
\end{APACrefauthors}%
\unskip\
\newblock
\APACrefYearMonthDay{1980}{}{}.
\newblock
{\BBOQ}\APACrefatitle {A semantics for positive and comparative adjectives} {A
  semantics for positive and comparative adjectives}.{\BBCQ}
\newblock
\APACjournalVolNumPages{Linguistics and philosophy}{4}{1}{1--45}.
\PrintBackRefs{\CurrentBib}

\bibitem [\protect \citeauthoryear {%
Kratzer%
\ \BBA {} Irene%
}{%
Kratzer%
\ \BBA {} Irene%
}{%
{\protect \APACyear {1998}}%
}]{%
kratzer1998semantics}
\APACinsertmetastar {%
kratzer1998semantics}%
\begin{APACrefauthors}%
Kratzer, A.%
\BCBT {}\ \BBA {} Irene, H.%
\end{APACrefauthors}%
\unskip\
\newblock
\APACrefYear{1998}.
\newblock
\APACrefbtitle {Semantics in generative grammar} {Semantics in generative
  grammar}\ (\BVOL\ 1185).
\newblock
\APACaddressPublisher{}{Blackwell Oxford}.
\PrintBackRefs{\CurrentBib}

\bibitem [\protect \citeauthoryear {%
Kripke%
}{%
Kripke%
}{%
{\protect \APACyear {1963}}%
}]{%
kripke1963semantical}
\APACinsertmetastar {%
kripke1963semantical}%
\begin{APACrefauthors}%
Kripke, S\BPBI A.%
\end{APACrefauthors}%
\unskip\
\newblock
\APACrefYearMonthDay{1963}{}{}.
\newblock
{\BBOQ}\APACrefatitle {Semantical analysis of modal logic i normal modal
  propositional calculi} {Semantical analysis of modal logic i normal modal
  propositional calculi}.{\BBCQ}
\newblock
\APACjournalVolNumPages{Mathematical Logic Quarterly}{9}{5-6}{67--96}.
\PrintBackRefs{\CurrentBib}

\bibitem [\protect \citeauthoryear {%
Kwiatkowski%
, Zettlemoyer%
, Goldwater%
\BCBL {}\ \BBA {} Steedman%
}{%
Kwiatkowski%
\ \protect \BOthers {.}}{%
{\protect \APACyear {2011}}%
}]{%
kwiatkowski2011lexical}
\APACinsertmetastar {%
kwiatkowski2011lexical}%
\begin{APACrefauthors}%
Kwiatkowski, T.%
, Zettlemoyer, L.%
, Goldwater, S.%
\BCBL {}\ \BBA {} Steedman, M.%
\end{APACrefauthors}%
\unskip\
\newblock
\APACrefYearMonthDay{2011}{}{}.
\newblock
{\BBOQ}\APACrefatitle {Lexical generalization in CCG grammar induction for
  semantic parsing} {Lexical generalization in ccg grammar induction for
  semantic parsing}.{\BBCQ}
\newblock
\BIn{} \APACrefbtitle {Proceedings of the 2011 Conference on Empirical Methods
  in Natural Language Processing} {Proceedings of the 2011 conference on
  empirical methods in natural language processing}\ (\BPGS\ 1512--1523).
\PrintBackRefs{\CurrentBib}

\bibitem [\protect \citeauthoryear {%
Lassiter%
\ \BBA {} Goodman%
}{%
Lassiter%
\ \BBA {} Goodman%
}{%
{\protect \APACyear {2017}}%
}]{%
lassiter2017adjectival}
\APACinsertmetastar {%
lassiter2017adjectival}%
\begin{APACrefauthors}%
Lassiter, D.%
\BCBT {}\ \BBA {} Goodman, N\BPBI D.%
\end{APACrefauthors}%
\unskip\
\newblock
\APACrefYearMonthDay{2017}{}{}.
\newblock
{\BBOQ}\APACrefatitle {Adjectival vagueness in a Bayesian model of
  interpretation} {Adjectival vagueness in a bayesian model of
  interpretation}.{\BBCQ}
\newblock
\APACjournalVolNumPages{Synthese}{194}{10}{3801--3836}.
\PrintBackRefs{\CurrentBib}

\bibitem [\protect \citeauthoryear {%
Levinson%
}{%
Levinson%
}{%
{\protect \APACyear {1983}}%
}]{%
levinson1983pragmatics}
\APACinsertmetastar {%
levinson1983pragmatics}%
\begin{APACrefauthors}%
Levinson, S\BPBI C.%
\end{APACrefauthors}%
\unskip\
\newblock
\APACrefYear{1983}.
\newblock
\APACrefbtitle {Pragmatics} {Pragmatics}.
\newblock
\APACaddressPublisher{}{Cambridge university press}.
\PrintBackRefs{\CurrentBib}

\bibitem [\protect \citeauthoryear {%
Lin%
}{%
Lin%
}{%
{\protect \APACyear {1991}}%
}]{%
lin1991divergence}
\APACinsertmetastar {%
lin1991divergence}%
\begin{APACrefauthors}%
Lin, J.%
\end{APACrefauthors}%
\unskip\
\newblock
\APACrefYearMonthDay{1991}{}{}.
\newblock
{\BBOQ}\APACrefatitle {Divergence measures based on the Shannon entropy}
  {Divergence measures based on the shannon entropy}.{\BBCQ}
\newblock
\APACjournalVolNumPages{IEEE Transactions on Information
  theory}{37}{1}{145--151}.
\PrintBackRefs{\CurrentBib}

\bibitem [\protect \citeauthoryear {%
Mishra%
\ \protect \BOthers {.}}{%
Mishra%
\ \protect \BOthers {.}}{%
{\protect \APACyear {2022}}%
}]{%
mishra2022lila}
\APACinsertmetastar {%
mishra2022lila}%
\begin{APACrefauthors}%
Mishra, S.%
, Finlayson, M.%
, Lu, P.%
, Tang, L.%
, Welleck, S.%
, Baral, C.%
\BDBL {}others%
\end{APACrefauthors}%
\unskip\
\newblock
\APACrefYearMonthDay{2022}{}{}.
\newblock
{\BBOQ}\APACrefatitle {Lila: A unified benchmark for mathematical reasoning}
  {Lila: A unified benchmark for mathematical reasoning}.{\BBCQ}
\newblock
\APACjournalVolNumPages{arXiv preprint arXiv:2210.17517}{}{}{}.
\PrintBackRefs{\CurrentBib}

\bibitem [\protect \citeauthoryear {%
Montague%
}{%
Montague%
}{%
{\protect \APACyear {1973}}%
}]{%
montague1973proper}
\APACinsertmetastar {%
montague1973proper}%
\begin{APACrefauthors}%
Montague, R.%
\end{APACrefauthors}%
\unskip\
\newblock
\APACrefYearMonthDay{1973}{}{}.
\newblock
{\BBOQ}\APACrefatitle {The proper treatment of quantification in ordinary
  English} {The proper treatment of quantification in ordinary english}.{\BBCQ}
\newblock
\BIn{} \APACrefbtitle {Approaches to natural language: Proceedings of the 1970
  Stanford workshop on grammar and semantics} {Approaches to natural language:
  Proceedings of the 1970 stanford workshop on grammar and semantics}\ (\BPGS\
  221--242).
\PrintBackRefs{\CurrentBib}

\bibitem [\protect \citeauthoryear {%
Nelder%
\ \BBA {} Mead%
}{%
Nelder%
\ \BBA {} Mead%
}{%
{\protect \APACyear {1965}}%
}]{%
nelder1965simplex}
\APACinsertmetastar {%
nelder1965simplex}%
\begin{APACrefauthors}%
Nelder, J\BPBI A.%
\BCBT {}\ \BBA {} Mead, R.%
\end{APACrefauthors}%
\unskip\
\newblock
\APACrefYearMonthDay{1965}{}{}.
\newblock
{\BBOQ}\APACrefatitle {A simplex method for function minimization} {A simplex
  method for function minimization}.{\BBCQ}
\newblock
\APACjournalVolNumPages{The computer journal}{7}{4}{308--313}.
\PrintBackRefs{\CurrentBib}

\bibitem [\protect \citeauthoryear {%
Partee%
, ter Meulen%
\BCBL {}\ \BBA {} Wall%
}{%
Partee%
\ \protect \BOthers {.}}{%
{\protect \APACyear {1990}}%
}]{%
partee1990mathematical}
\APACinsertmetastar {%
partee1990mathematical}%
\begin{APACrefauthors}%
Partee, B\BPBI B.%
, ter Meulen, A.%
\BCBL {}\ \BBA {} Wall, R.%
\end{APACrefauthors}%
\unskip\
\newblock
\APACrefYear{1990}.
\newblock
\APACrefbtitle {Mathematical Methods in Linguistics} {Mathematical methods in
  linguistics}\ (\BVOL~30).
\newblock
\APACaddressPublisher{}{Springer Science \& Business Media}.
\PrintBackRefs{\CurrentBib}

\bibitem [\protect \citeauthoryear {%
Piantadosi%
}{%
Piantadosi%
}{%
{\protect \APACyear {2023}}%
}]{%
piantadosi2023modern}
\APACinsertmetastar {%
piantadosi2023modern}%
\begin{APACrefauthors}%
Piantadosi, S\BPBI T.%
\end{APACrefauthors}%
\unskip\
\newblock
\APACrefYearMonthDay{2023}{}{}.
\newblock
{\BBOQ}\APACrefatitle {Modern language models refute Chomsky’s approach to
  language} {Modern language models refute chomsky’s approach to
  language}.{\BBCQ}
\newblock
\APACjournalVolNumPages{Lingbuzz Preprint, lingbuzz/007180}{}{}{}.
\PrintBackRefs{\CurrentBib}

\bibitem [\protect \citeauthoryear {%
Qing%
\ \BBA {} Franke%
}{%
Qing%
\ \BBA {} Franke%
}{%
{\protect \APACyear {2014}}%
}]{%
qing2014gradable}
\APACinsertmetastar {%
qing2014gradable}%
\begin{APACrefauthors}%
Qing, C.%
\BCBT {}\ \BBA {} Franke, M.%
\end{APACrefauthors}%
\unskip\
\newblock
\APACrefYearMonthDay{2014}{}{}.
\newblock
{\BBOQ}\APACrefatitle {Gradable adjectives, vagueness, and optimal language
  use: A speaker-oriented model} {Gradable adjectives, vagueness, and optimal
  language use: A speaker-oriented model}.{\BBCQ}
\newblock
\BIn{} \APACrefbtitle {Semantics and linguistic theory} {Semantics and
  linguistic theory}\ (\BVOL~24, \BPGS\ 23--41).
\PrintBackRefs{\CurrentBib}

\bibitem [\protect \citeauthoryear {%
Ruis%
\ \protect \BOthers {.}}{%
Ruis%
\ \protect \BOthers {.}}{%
{\protect \APACyear {2022}}%
}]{%
ruis2022large}
\APACinsertmetastar {%
ruis2022large}%
\begin{APACrefauthors}%
Ruis, L.%
, Khan, A.%
, Biderman, S.%
, Hooker, S.%
, Rockt{\"a}schel, T.%
\BCBL {}\ \BBA {} Grefenstette, E.%
\end{APACrefauthors}%
\unskip\
\newblock
\APACrefYearMonthDay{2022}{}{}.
\newblock
{\BBOQ}\APACrefatitle {Large language models are not zero-shot communicators}
  {Large language models are not zero-shot communicators}.{\BBCQ}
\newblock
\APACjournalVolNumPages{arXiv preprint arXiv:2210.14986}{}{}{}.
\PrintBackRefs{\CurrentBib}

\bibitem [\protect \citeauthoryear {%
Searle%
}{%
Searle%
}{%
{\protect \APACyear {1969}}%
}]{%
searle1969speech}
\APACinsertmetastar {%
searle1969speech}%
\begin{APACrefauthors}%
Searle, J\BPBI R.%
\end{APACrefauthors}%
\unskip\
\newblock
\APACrefYear{1969}.
\newblock
\APACrefbtitle {Speech acts: An essay in the philosophy of language} {Speech
  acts: An essay in the philosophy of language}\ (\BVOL~626).
\newblock
\APACaddressPublisher{}{Cambridge university press}.
\PrintBackRefs{\CurrentBib}

\bibitem [\protect \citeauthoryear {%
Steedman%
}{%
Steedman%
}{%
{\protect \APACyear {2001}}%
}]{%
steedman2001syntactic}
\APACinsertmetastar {%
steedman2001syntactic}%
\begin{APACrefauthors}%
Steedman, M.%
\end{APACrefauthors}%
\unskip\
\newblock
\APACrefYear{2001}.
\newblock
\APACrefbtitle {The syntactic process} {The syntactic process}.
\newblock
\APACaddressPublisher{}{MIT press}.
\PrintBackRefs{\CurrentBib}

\bibitem [\protect \citeauthoryear {%
Tenney%
, Das%
\BCBL {}\ \BBA {} Pavlick%
}{%
Tenney%
\ \protect \BOthers {.}}{%
{\protect \APACyear {2019}}%
}]{%
tenney2019bert}
\APACinsertmetastar {%
tenney2019bert}%
\begin{APACrefauthors}%
Tenney, I.%
, Das, D.%
\BCBL {}\ \BBA {} Pavlick, E.%
\end{APACrefauthors}%
\unskip\
\newblock
\APACrefYearMonthDay{2019}{}{}.
\newblock
{\BBOQ}\APACrefatitle {BERT rediscovers the classical NLP pipeline} {Bert
  rediscovers the classical nlp pipeline}.{\BBCQ}
\newblock
\APACjournalVolNumPages{arXiv preprint arXiv:1905.05950}{}{}{}.
\PrintBackRefs{\CurrentBib}

\bibitem [\protect \citeauthoryear {%
Tessler%
\ \BBA {} Goodman%
}{%
Tessler%
\ \BBA {} Goodman%
}{%
{\protect \APACyear {2022}}%
}]{%
tessler2022warm}
\APACinsertmetastar {%
tessler2022warm}%
\begin{APACrefauthors}%
Tessler, M\BPBI H.%
\BCBT {}\ \BBA {} Goodman, N\BPBI D.%
\end{APACrefauthors}%
\unskip\
\newblock
\APACrefYearMonthDay{2022}{}{}.
\newblock
{\BBOQ}\APACrefatitle {Warm (for Winter): Inferring comparison classes in
  communication} {Warm (for winter): Inferring comparison classes in
  communication}.{\BBCQ}
\newblock
\APACjournalVolNumPages{Cognitive Science}{46}{3}{e13095}.
\PrintBackRefs{\CurrentBib}

\bibitem [\protect \citeauthoryear {%
Tessler%
, Tsvilodub%
, Snedeker%
\BCBL {}\ \BBA {} Levy%
}{%
Tessler%
\ \protect \BOthers {.}}{%
{\protect \APACyear {2020}}%
}]{%
tessler2020informational}
\APACinsertmetastar {%
tessler2020informational}%
\begin{APACrefauthors}%
Tessler, M\BPBI H.%
, Tsvilodub, P.%
, Snedeker, J.%
\BCBL {}\ \BBA {} Levy, R\BPBI P.%
\end{APACrefauthors}%
\unskip\
\newblock
\APACrefYearMonthDay{2020}{}{}.
\newblock
{\BBOQ}\APACrefatitle {Informational goals, sentence structure, and comparison
  class inference} {Informational goals, sentence structure, and comparison
  class inference}.{\BBCQ}
\newblock
\BIn{} \APACrefbtitle {Proceedings of the Annual Conference of the Cognitive
  Science Society.} {Proceedings of the annual conference of the cognitive
  science society.}
\PrintBackRefs{\CurrentBib}

\bibitem [\protect \citeauthoryear {%
Van~Eijck%
\ \BBA {} Lappin%
}{%
Van~Eijck%
\ \BBA {} Lappin%
}{%
{\protect \APACyear {2012}}%
}]{%
van2012probabilistic}
\APACinsertmetastar {%
van2012probabilistic}%
\begin{APACrefauthors}%
Van~Eijck, J.%
\BCBT {}\ \BBA {} Lappin, S.%
\end{APACrefauthors}%
\unskip\
\newblock
\APACrefYearMonthDay{2012}{}{}.
\newblock
{\BBOQ}\APACrefatitle {Probabilistic semantics for natural language}
  {Probabilistic semantics for natural language}.{\BBCQ}
\newblock
\APACjournalVolNumPages{Logic and interactive rationality (LIRA)}{2}{}{17--35}.
\PrintBackRefs{\CurrentBib}

\bibitem [\protect \citeauthoryear {%
White%
, Mu%
\BCBL {}\ \BBA {} Goodman%
}{%
White%
\ \protect \BOthers {.}}{%
{\protect \APACyear {2020}}%
}]{%
white2020learning}
\APACinsertmetastar {%
white2020learning}%
\begin{APACrefauthors}%
White, J.%
, Mu, J.%
\BCBL {}\ \BBA {} Goodman, N\BPBI D.%
\end{APACrefauthors}%
\unskip\
\newblock
\APACrefYearMonthDay{2020}{}{}.
\newblock
{\BBOQ}\APACrefatitle {Learning to refer informatively by amortizing pragmatic
  reasoning} {Learning to refer informatively by amortizing pragmatic
  reasoning}.{\BBCQ}
\newblock
\APACjournalVolNumPages{arXiv preprint arXiv:2006.00418}{}{}{}.
\PrintBackRefs{\CurrentBib}

\bibitem [\protect \citeauthoryear {%
Wittgenstein%
}{%
Wittgenstein%
}{%
{\protect \APACyear {1953}}%
}]{%
wittgenstein2010philosophical}
\APACinsertmetastar {%
wittgenstein2010philosophical}%
\begin{APACrefauthors}%
Wittgenstein, L.%
\end{APACrefauthors}%
\unskip\
\newblock
\APACrefYear{1953}.
\newblock
\APACrefbtitle {Philosophical investigations} {Philosophical investigations}.
\PrintBackRefs{\CurrentBib}

\bibitem [\protect \citeauthoryear {%
Wong%
\ \protect \BOthers {.}}{%
Wong%
\ \protect \BOthers {.}}{%
{\protect \APACyear {prep.}}%
}]{%
wong2023translating}
\APACinsertmetastar {%
wong2023translating}%
\begin{APACrefauthors}%
Wong, L.%
, Grand, G.%
, Lew, A.%
, Andreas, J.%
, Goodman, N\BPBI D.%
, Mansinghka, V.%
\BCBL {}\ \BBA {} Tenenbaum, J\BPBI B.%
\end{APACrefauthors}%
\unskip\
\newblock
\APACrefYearMonthDay{prep.}{}{}.
\newblock
{\BBOQ}\APACrefatitle {Translating from Natural Language to the Language of
  Thought} {Translating from natural language to the language of
  thought}.{\BBCQ}
\newblock
\APACjournalVolNumPages{preprint}{}{}{}.
\PrintBackRefs{\CurrentBib}

\bibitem [\protect \citeauthoryear {%
Wu%
\ \BBA {} Goodman%
}{%
Wu%
\ \BBA {} Goodman%
}{%
{\protect \APACyear {2022}}%
}]{%
wu2022foundation}
\APACinsertmetastar {%
wu2022foundation}%
\begin{APACrefauthors}%
Wu, M.%
\BCBT {}\ \BBA {} Goodman, N.%
\end{APACrefauthors}%
\unskip\
\newblock
\APACrefYearMonthDay{2022}{}{}.
\newblock
{\BBOQ}\APACrefatitle {Foundation Posteriors for Approximate Probabilistic
  Inference} {Foundation posteriors for approximate probabilistic
  inference}.{\BBCQ}
\newblock
\APACjournalVolNumPages{arXiv preprint arXiv:2205.09735}{}{}{}.
\PrintBackRefs{\CurrentBib}

\bibitem [\protect \citeauthoryear {%
Zelikman%
, Huang%
, Poesia%
, Goodman%
\BCBL {}\ \BBA {} Haber%
}{%
Zelikman%
\ \protect \BOthers {.}}{%
{\protect \APACyear {2022}}%
}]{%
zelikman2022parsel}
\APACinsertmetastar {%
zelikman2022parsel}%
\begin{APACrefauthors}%
Zelikman, E.%
, Huang, Q.%
, Poesia, G.%
, Goodman, N\BPBI D.%
\BCBL {}\ \BBA {} Haber, N.%
\end{APACrefauthors}%
\unskip\
\newblock
\APACrefYearMonthDay{2022}{}{}.
\newblock
{\BBOQ}\APACrefatitle {Parsel: A Unified Natural Language Framework for
  Algorithmic Reasoning} {Parsel: A unified natural language framework for
  algorithmic reasoning}.{\BBCQ}
\newblock
\APACjournalVolNumPages{arXiv preprint arXiv:2212.10561}{}{}{}.
\PrintBackRefs{\CurrentBib}

\end{thebibliography}

\end{document}